\documentclass{article}
\usepackage[margin=1in]{geometry}
\usepackage{graphicx}
\usepackage{booktabs}
\usepackage{orcidlink}
\usepackage{subcaption}
\usepackage{CJKutf8}
\usepackage{pifont}
\usepackage{amsmath,amssymb}
\usepackage{yhmath}
\usepackage{xcolor}
\usepackage{booktabs}
\usepackage{makecell}
\usepackage[table]{xcolor}
\newcommand\scalemath[2]{\scalebox{#1}{\mbox{\ensuremath{\displaystyle #2}}}}

\DeclareMathAlphabet{\bbold}{U}{bbold}{m}{n}

\usepackage{multirow}
\usepackage{algorithm}
\usepackage{algpseudocode}
\algtext*{EndWhile}
\algtext*{EndIf}
\algtext*{EndFor}
\usepackage{array}
\newcommand{\ourbench}{{{360Bench}}}
\newcommand{\ourmodel}{{{Free360}}}
\newcommand{\degree}{{{°}}}

\definecolor{profcolor}{RGB}{255, 102, 102}   \definecolor{quangcolor}{RGB}{153, 102, 51}  \definecolor{trancolor}{RGB}{0, 0, 255}    
\definecolor{navy}{RGB}{50,110,168}
\definecolor{darkgreen}{RGB}{50,110,225}
\definecolor{lightorange}{RGB}{255,200,120}
\definecolor{timecolor}{RGB}{255, 102, 102}

\usepackage[most]{tcolorbox} 
\usepackage{enumitem} 

\newcommand{\qt}[1]{\textcolor{black}{#1}} 
\newcommand{\trans}[1]{\textcolor{black}{#1}} 
\newcommand{\tran}[1]{\textcolor{black}{#1}} 
\newcommand{\kj}[1]{\textcolor{black}{#1}} 
\newcommand{\ot}[1]{\textcolor{black}{#1}}

\newcommand{\imt}[1]{\textcolor{darkgreen}{#1}} 

\date{}
\begin{document}

\title{360\degree{} Image Perception with MLLMs: \\A Comprehensive Benchmark and \\a Training-Free Method} 

\author{
Huyen T. T. Tran$^{1}$\thanks{Corresponding author: tranhuyen1191@gmail.com}, 
Van-Quang Nguyen$^{2}$, 
Farros Alferro$^{1}$, \\
Kang-Jun Liu$^{1}$, 
Takayuki Okatani$^{1,2}$ \\
\\
$^{1}$ GSIS, Tohoku University \\
$^{2}$ RIKEN AIP, Japan \\
}

\maketitle

\begin{abstract}

  Multimodal Large Language Models (MLLMs) have shown impressive abilities in understanding and reasoning over conventional images. However, their perception of 360\degree{} images remains largely underexplored. Unlike conventional images, 360\degree{} images capture the entire surrounding environment, enabling holistic spatial reasoning but introducing challenges such as geometric distortion and complex spatial relations. To comprehensively assess MLLMs' capabilities to perceive 360\degree{} images, we introduce \textbf{360Bench}, a Visual Question Answering (VQA) benchmark featuring 7K-resolution 360\degree{} images, seven representative (sub)tasks with annotations carefully curated by human annotators. Using 360Bench, we systematically evaluate seven MLLMs and six enhancement methods, revealing their shortcomings in 360\degree{} image perception. To address these challenges, we propose \textbf{Free360}, a training-free scene-graph-based framework for high-resolution 360$^\circ$ VQA.
  Free360 decomposes the reasoning process into modular steps, applies adaptive spherical image transformations to 360\degree{} images tailored to each step, and seamlessly integrates the resulting information into a unified graph representation for answer generation. 
  Experiments show that Free360 consistently improves its base MLLM and provides a strong training-free solution for 360$^\circ$ VQA tasks. The source code and dataset will be publicly released at https://github.com/TranHuyen1191/360Bench-Free360.
\end{abstract}

\section{Introduction}\label{sec:Introduction}

The development of MLLMs, from proprietary to open-source models \cite{GPT4o,gemini25,InternVL3, qwen25vltechnicalreport},
has enabled models to perform complex tasks involving the understanding and integration of multimodal information \cite{yin2024survey,chang2024survey}. In contrast to conventional models, which are typically trained on small datasets tailored to specific tasks (e.g., image captioning), MLLMs are pre-trained on massive corpora, allowing them to tackle a wide range of tasks via natural language instructions \cite{m6,Kosmos1}. As a result, MLLMs 
\qt{have been increasingly} 
applied in diverse real-world domains, such as healthcare, robotics, and education~\cite{li2024surveyingmllmlandscapemetareview,MLLMforEdu}.

Unlike conventional images, 360\degree{} images offer a complete view of the surrounding environment within a single frame, enabling comprehensive spatial reasoning across the entire scene---a capability that single or even multiple conventional images cannot effectively provide. \ot{While still uncommon, enabling MLLMs to handle 360° images holds great promise for applications requiring holistic scene understanding, such as autonomous driving, assistive robotics, and surveillance.}

\begin{figure}[!ht]
\centering
\includegraphics[width=\columnwidth]{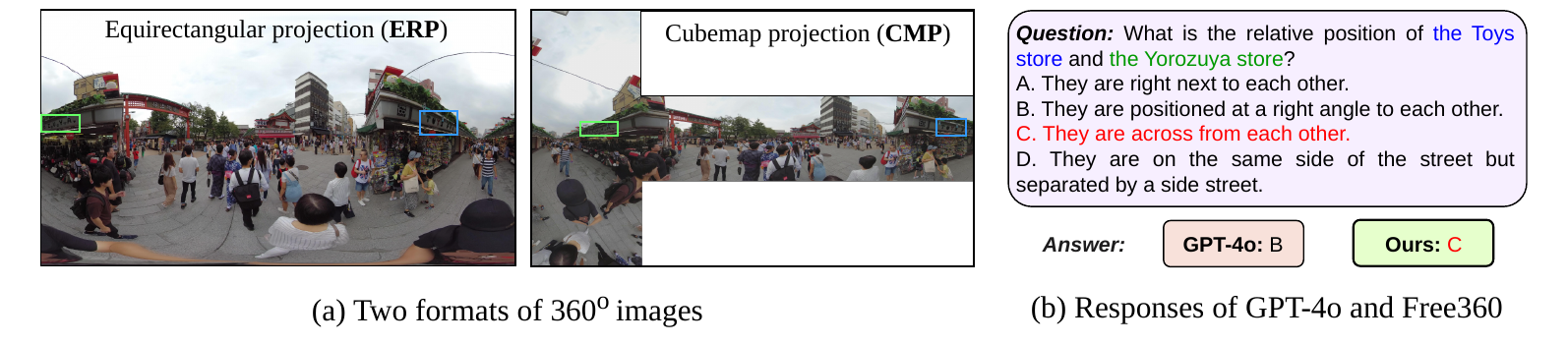}
\caption{\qt{(a)  Example of an image from \textbf{\ourbench{}} in two common projection formats with relevant entities highlighted. (b) Responses of GPT-4o and \textbf{\ourmodel{}} (Ours) with correct answers highlighted in red. GPT-4o struggles with spatial reasoning, whereas \ourmodel{} correctly infers spatial relations. 360\degree{} image source:~\cite{Insta360}.}
}
\label{fig:Ilustration}
\end{figure}

\ot{However, realizing this potential in MLLMs comes with significant challenges.}
First, {\textit{spatial reasoning}} over a continuous spherical space is inherently complex. 
\qt{As shown in Fig.~\ref{fig:Ilustration}(b), MLLMs often misinterpret spatial relations. Second, 360\degree{} images have the extensive field of view with \tran{rich} visual content, requiring models to capture \textit{fine-grained features} for accurate understanding.}
\qt{Third, since most models are designed for planar images, 360\degree{} images are usually projected into \tran{planar representations} rather than processed in their native spherical form. 
Such projections cause \textit{geometric distortion} (e.g., stretched poles from uneven sampling) and \textit{object fragmentation} (\tran{when objects span across image} boundaries).}
\qt{Among projection methods, \textit{EquiRectangular Projection (ERP)} and \textit{CubeMap Projection (CMP)} are the most commonly used~\cite{ERPmostused,nguyen2019optimal}, as illustrated in Fig.~\ref{fig:Ilustration}(a).}
\qt{ERP maps a sphere's longitude and latitude to 2D horizontal and vertical coordinates~\cite{rao2021omnilayout}, whereas CMP projects the sphere onto six cube faces~\cite{cheng2018cube,jung2025edm}.}
To date, it remains unclear \tran{which format facilitates MLLM perception more effectively.
} 

{\color{black}
Recent efforts have begun to explore question answering and reasoning over omnidirectional data~\cite{omniVQA,odibench}. OmniVQA~\cite{omniVQA} introduces a 360$^\circ$ VQA dataset built from 1440$\times$720 ERP images of indoor scenes and evaluates five MLLM families on two tasks derived from six templates. Very recently, concurrent work has also started to investigate broader 360$^\circ$ image understanding. In particular, ODI-Bench~\cite{odibench} studies ten VQA tasks and introduces a training-free reasoning framework for MLLMs. We view this line of work as complementary to ours. In contrast, our focus is high-resolution  360$^\circ$ VQA with manually curated annotations and subtasks designed to probe fine-grained perception, spatial reasoning, and robustness to projection-induced challenges.
}

To bridge this gap, we introduce \textbf{\ourbench{}}, a 360\degree{} VQA benchmark with 7K-resolution images spanning indoor, outdoor, and aerial scenes. \ourbench{} comprises 1,532 unique samples featuring curated annotations across seven (sub)-tasks. 
Table~\ref{tab:dataset_comparison} compares \ourbench{} with existing VQA datasets in terms of image resolution, scene diversity, annotation methods, and \qt{task scopes}. Using 360Bench, we systematically evaluate seven MLLMs and six enhancement methods, \qt{revealing notable performance gaps: the best model achieves only 46.5\% accuracy, far below human performance} of 86.3\%. 
\qt{Regarding the projection format, CMP shows a clear advantage on projection-distorted tasks (up to 14.1\%), while ERP generally performs better on spatial reasoning tasks (up to 14.6\%), underscoring their complementary strengths.}

\begin{table*}[t]
\centering
\caption{Comparison between \ourbench{} and existing VQA datasets. 
\textbf{CI}, \textbf{360I},  \textbf{360V}, and \textbf{c360I} denote conventional images, 360\degree{} images, 360\degree{} videos, and cross-frame correlated 360\degree{} images, respectively.
\textbf{Res.} refers to image resolution. 
\textbf{Temp. (\#)} indicates template-based annotations, with the number of templates in parentheses. 
\textbf{Auto} refers to annotations generated automatically using vision models. 
\textbf{Manual} denotes human-crafted annotations. 
\textbf{\#Samples (\#)}  indicates the number of unique samples, with test samples shown in parentheses. 
\textbf{Avail.} indicates whether the dataset is publicly accessible \trans{at the time of writing.}}
\label{tab:dataset_comparison}
\renewcommand{\arraystretch}{1.1}
\resizebox{\textwidth}{!}{%
\begin{tabular}{l|cccccccccccccc}
\Xhline{1.2pt}
\textbf{Dataset} & \textbf{Type} & \textbf{Res.} & \textbf{Scene} & \textbf{Annotation} & \textbf{\#Samples} & \textbf{\#Tasks} & \textbf{FP-IR} &\textbf{FP-IC}  & \textbf{PP-IR} & \textbf{PP-IC} & \textbf{SR-Os} & \textbf{SR-OV} & \textbf{DG}& \textbf{Avail.} \\
\hline
 HR-Bench~\cite{DC2}         & CI         & 8K     & Diverse         & Manual         &200 (200)&2& \ding{51}   & \ding{55}    & \ding{55}       & \ding{55}      & \ding{51} & \ding{55}      & \ding{55} & \ding{55} \\ 
{Pano-AVQA}~\cite{PanoAVQA}       & 360V         & N/A                 & Diverse        & Temp. (7)          &51.7K (5.3K)&2& \ding{55}   & \ding{55}    & \ding{55}       & \ding{55}      & \ding{51} & \ding{55}      & \ding{55}& \ding{55} \\ 
{CFpano}~\cite{CFpano}         & c360I         &  1K    & Indoor         & Temp. (36)          &8094 (1619)&17& \ding{55}   & \ding{55}    & \ding{55}       & \ding{55}      & \ding{55} & \ding{55}      & \ding{55}& \ding{55} \\
{VQA360}~\cite{VQA360}          & 360I         & 1K     & Indoor         & Temp. (17)         &16,945 (6962)&3& \ding{55}   & \ding{55}    & \ding{55}       & \ding{55}      & \ding{51} & \ding{51}      & \ding{55} & \ding{55} \\ 
{OmniVQA}~\cite{omniVQA}         & 360I         & 1K     & Indoor         & Temp. (6)          &5652 (800)&2& \ding{55}   & \ding{55}    & \ding{51}       & \ding{55}      & \ding{51} & \ding{55}      & \ding{55}& \ding{55} \\
{ODI-Bench}~\cite{odibench}         & 360I         & Mixed     & Diverse         & Auto\&Manual          &4254 (4254)&10& \ding{55}   & \ding{55}    & \ding{51}       & \ding{55}      & \ding{51} & \ding{51}      & \ding{55}& \ding{55} \\
\rowcolor{blue!10} \textbf{\ourbench{} (Ours)}   & 360I         & 7K    & Diverse        & Manual           &1532 (1532)&7 & \ding{51}   & \ding{51}    & \ding{51}       & \ding{51}     & \ding{51}  & \ding{51}      & \ding{51} & \ding{51}\\
\Xhline{1.2pt}
\end{tabular}
}
\end{table*}

\qt{
These results highlight the fundamental limitations of existing MLLMs in understanding 360\degree{} scenes. While fine-tuning could improve performance, it is computationally expensive, labor-intensive, and must be repeated for each downstream
task~\cite{long2024trainingfreeunsupervisedpromptvisionlanguage,he2026tensor}. 
Moreover, fine-tuning risks catastrophic forgetting, erasing general knowledge acquired during pretraining~\cite{luo2025empiricalstudycatastrophicforgetting,liao2025exploring,zhai2024investigating}. In contrast, training-free approaches offer a scalable and practical alternative that preserves pretrained knowledge, crucial for large-scale MLLMs.
}

{\color{black}
To address the challenges of 360$^\circ$ image perception, we propose \textbf{\ourmodel}, a training-free scene-graph-based framework for high-resolution single-image 360$^\circ$ VQA.
}
The use of scene graphs enables (1) decomposition of the reasoning process into modular steps, (2) application of adaptive spherical image transformations tailored to each step, and (3) seamless integration of the resulting information into a unified graph representation for final answer generation. Specifically, the scene graph generation (SGG) proceeds in four steps: (i) detecting question-relevant entities as nodes, (ii) extracting fine-grained attributes from cropped entity regions, (iii) inferring inter-entity relations via spherical rotations, and (iv) incorporating spatial relations between entities and the viewer. The resulting scene graph is then serialized into structured text and fed to the MLLM for question answering.
Experiments show that \ourmodel{} substantially {\color{black} improves the performance of its base MLLM}, yielding gains of up to 22.9\% on subtasks and 7.3\% overall, showing its effectiveness in bridging the gap between human and model perception of 360\degree{} scenes.

\qt{
Our contributions are summarized as follows:
\begin{itemize}
{\color{black}
    \item We introduce 360Bench, a high-resolution benchmark for single-image 360\degree{} VQA, comprising seven subtasks designed to evaluate fine-grained perception, spatial reasoning, and robustness to projection-induced challenges.}
    \item We systematically evaluate thirteen models, including MLLMs and enhanced variants, under both ERP and CMP settings on \ourbench{}, revealing their limitations in 360\degree{} image perception.
 {\color{black}
    \item We propose \ourmodel{}, a training-free scene-graph-based framework for high-resolution single-image 360\degree{} VQA that incorporates 360\degree{}-specific operations, including entity-centered rotation and view mapping, and consistently improves the performance of its base MLLM.
    }
\end{itemize}
}

\section{Related Work}\label{sec:RelatedWork}
\subsection{360\degree{} VQA Datasets}
{\color{black}
Several benchmarks have been developed for question answering and reasoning on omnidirectional data~\cite{VQA360, PanoAVQA,omniVQA,odibench,CFpano}. VQA360 \cite{VQA360} focuses on spatial reasoning between objects and between objects and the viewer in 360\degree{} images. Pano-AVQA~\cite{PanoAVQA} studies grounded question answering in 360\degree{} videos. OmniVQA~\cite{omniVQA} extends 360\degree{} image VQA to indoor ERP scenes with tasks centered on object recognition and spatial reasoning under geometric distortion. Related panoramic reasoning settings have also been explored beyond single-image VQA. For example, CFpano \cite{CFpano} considers cross-frame correlated panoramas, where reasoning is performed over multiple related panoramic observations rather than a single 360° image. In addition, a very recent concurrent benchmark, ODI-Bench~\cite{odibench}, broadens the problem from VQA to more general omnidirectional image understanding with a wider set of tasks.

Compared with these efforts, our benchmark is designed specifically for high-resolution single-image 360\degree{} VQA with manual annotations and seven subtasks targeting fine-grained perception, spatial reasoning, and robustness to projection-induced challenges. Thus, 360Bench complements existing omnidirectional benchmarks by emphasizing the setting where a single high-resolution panoramic image must support both local perception and global reasoning.
}


\subsection{MLLM Enhancement Methods}
{\color{black}
Despite recent advances, state-of-the-art MLLMs still show clear limitations in fine-grained perception and complex reasoning. Existing enhancement methods for conventional images, such as SEAL~\cite{SEAL}, DC$^2$~\cite{DC2}, and ZoomEye~\cite{zoomeye}, typically rely on patch-wise search or retrieval to identify informative regions and use them to assist downstream reasoning.

For omnidirectional images, dedicated enhancement strategies have only recently begun to emerge. The work in~\cite{omniVQA} proposes a rule-based reinforcement learning approach to fine-tune Qwen2.5-VL for 360\degree{} reasoning. Concurrent work \cite{odibench} introduces Omni-CoT, a training-free reasoning framework for omnidirectional image understanding. \trans{Given a question, Omni-CoT first generates an initial answer from textual descriptions of six perspective views, and subsequently refines the prediction using crops of relevant objects.}

Our method is complementary to these efforts. Rather than relying on additional training, we focus on high-resolution single-image 360\degree{} VQA and introduce a training-free framework that constructs question-relevant scene graphs through modular reasoning steps. This design enables the use of 360\degree{}-specific operations, such as entity-centered rotation and view mapping, to better model fine-grained attributes and spatial relations.
}

\subsection{MLLM-based Scene Graph Generation}
Recent studies have employed MLLMs for SGG~\cite{gpt4sgg,xu2025llava,elskhawy2025prism}. To mitigate the substantial cost of manual annotation, image caption datasets are frequently leveraged to automatically generate SGG datasets for training models~\cite{llm4sgg,ye2021linguistic,li2022integrating,shi2021simple}. Alternatively, \cite{elskhawy2025prism} introduces a training-free approach in which an LLM-based scene graph parser constructs scene graphs directly from image captions generated by MLLMs. Despite these advances, no prior work has explored using MLLMs for question-relevant scene graph generation, in which graphs are constructed specifically for each question.


\begin{figure}[t]
\centering
  \begin{subcaptionbox}{\label{fig:Breakdowntask}}[0.25\linewidth]
    {\includegraphics[width=\linewidth]{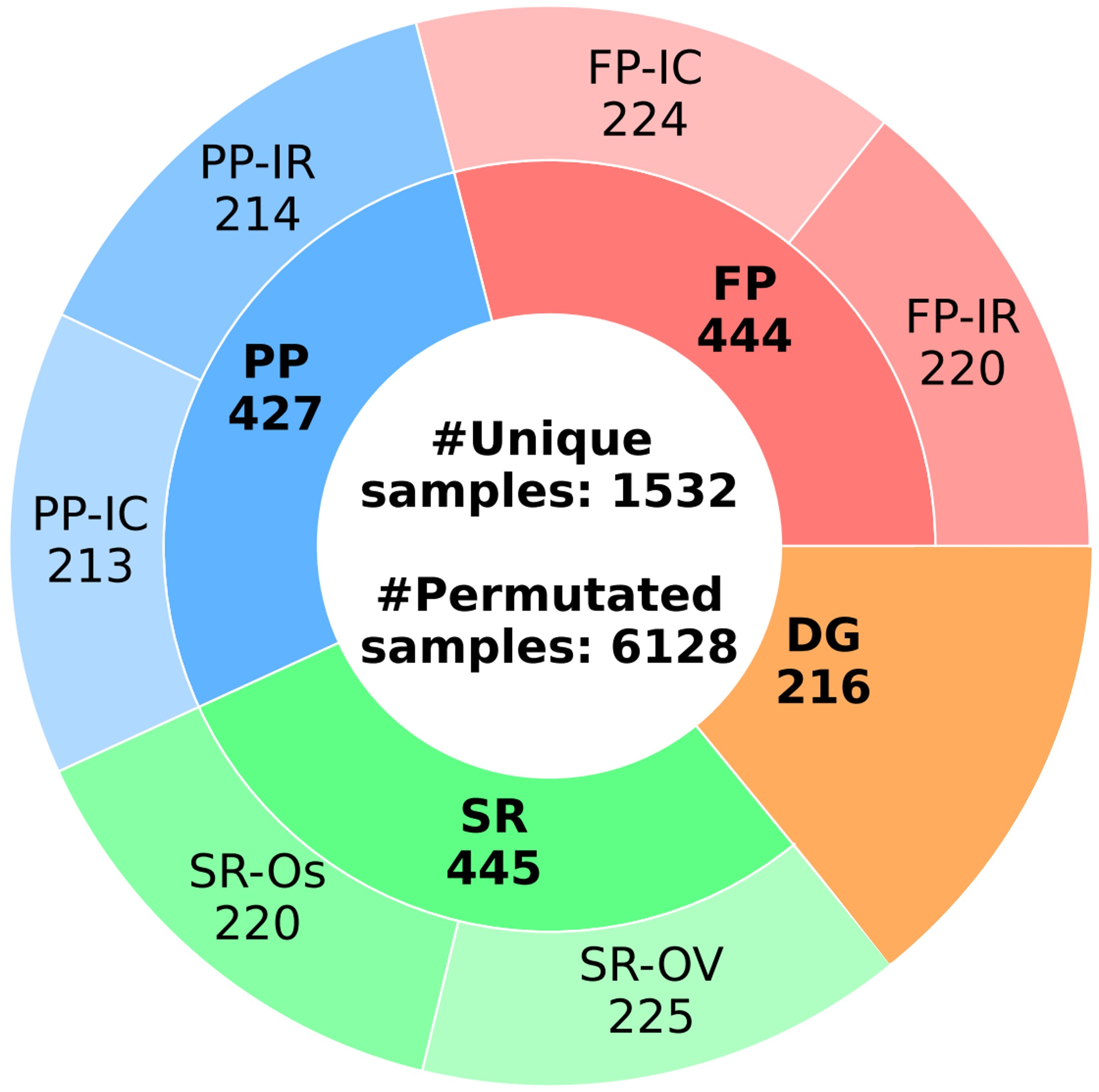}}
  \end{subcaptionbox}
  \begin{subcaptionbox}{\label{fig:category_distribution}}[0.35\linewidth]
    {\includegraphics[width=\linewidth]{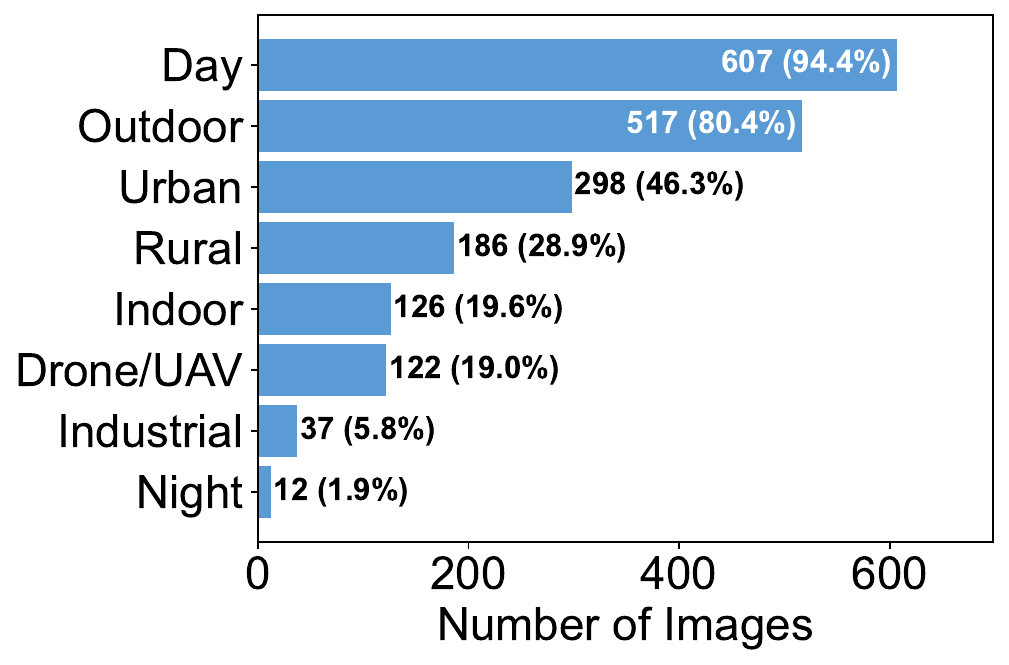}}
  \end{subcaptionbox}
  \begin{subcaptionbox}{\label{fig:bbox_analysis_scatter}}[0.35\linewidth]
    {\includegraphics[width=\linewidth]{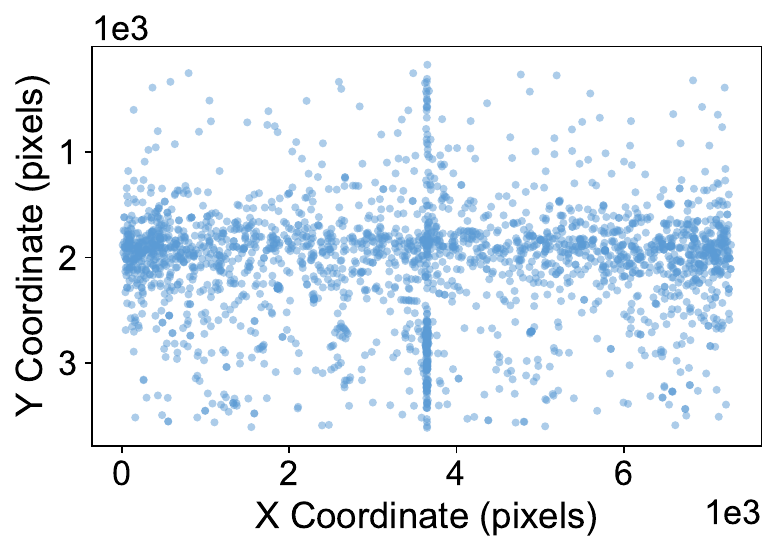}}    
  \end{subcaptionbox}
\caption{Statistics of \ourbench{}: (a) Task breakdown, (b) Image distribution, (c) Spatial distribution of annotated bounding boxes.}\label{fig:dataset_statistic}
\end{figure}

\section{\ourbench{} Benchmark}\label{sec:Dataset}
\begin{figure*}[t]
\centering
\includegraphics[width=\textwidth]{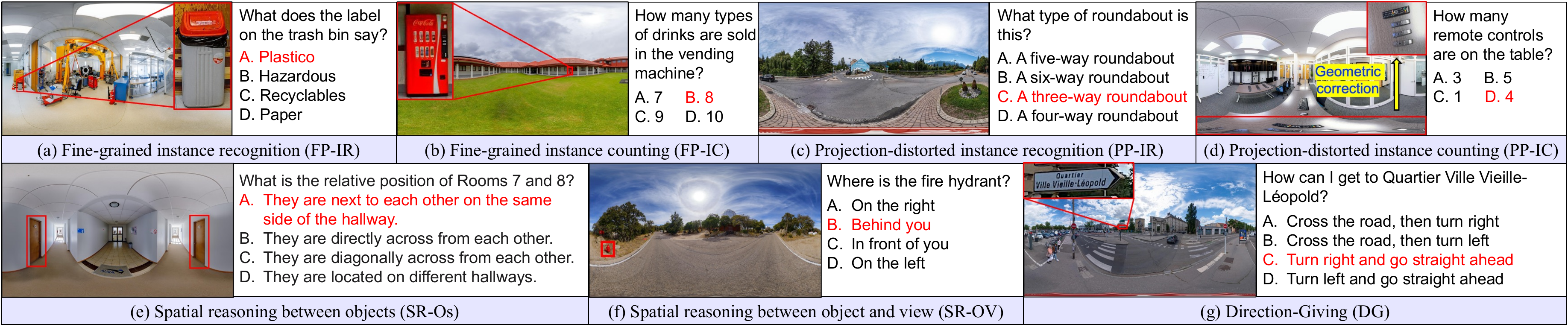} 
\caption{Illustration of tasks included in \ourbench{}. Correct answers \qt{and relevant entities} are highlighted in red. 360\degree{} image sources: \cite{NOIRLab, FlickHape}. }
\label{fig:QuestionExample}
\end{figure*}

\subsection{Task Definition}
\ourbench{} consists of four task categories: Fine-grained Perception (FP), Projec-tion-distorted Perception (PP), Spatial Reasoning (SR), and Direction-Giving (DG), as summarized in Fig.~\ref{fig:Breakdowntask}. Each task targets a distinct aspect of multimodal understanding under panoramic settings.

\qt{The Fine-grained Perception (FP) task evaluates a model's ability to recognize and reason about small or visually subtle instances, such as objects and text, within complex 360\degree{} scenes. 
It includes two subtasks: Instance Recognition (FP-IR), which focuses on identifying instance attributes, as illustrated in Fig.~\ref{fig:QuestionExample}(a), and Instance Counting (FP-IC), which measures the accurate enumeration of target objects, as shown in Fig.~\ref{fig:QuestionExample}(b).
}

\qt{The Projection-distorted Perception (PP) task assesses a model's robustness to geometric distortion and object fragmentation inherent in 360\degree{} projections. It contains two subtasks: Instance Recognition (PP-IR), which queries the attributes of distorted or split objects, and Instance Counting (PP-IC), which evaluates counting accuracy under severe distortions. To construct this task, we selected and annotated images with boundary-fragmented or heavily distorted objects. For instance, Fig.~\ref{fig:QuestionExample}(c) shows a PP-IR example where a road divided across the left and right boundaries appears as multiple paths, while Fig.~\ref{fig:QuestionExample}(d) depicts a PP-IC example where elongated remote controls complicate counting. 
}

\qt{The Spatial Reasoning (SR) task evaluates a model's ability to infer spatial relations in 360\degree{} scenes. It comprises 2 subtasks: Object-to-Object (SR-Os) reasoning, which examines relative positioning among objects (see Fig.~\ref{fig:QuestionExample}(e)), and Object-to-Viewer (SR-OV) reasoning, which focuses on the spatial relation between objects and the viewer (see Fig.~\ref{fig:QuestionExample}(f)). 
}

The Direction-Giving (DG) task tests goal-directed visual reasoning, requiring models to interpret spatial cues and plan multi-step routes toward a target, either visible or indicated by directional signs (see Fig.~\ref{fig:QuestionExample}(g)). Beyond SR, DG requires multi-step route planning that relies on holistic scene understanding. 

\qt{It is noted that for both the SR-OV and DG tasks, the viewer is assumed to be facing the direction corresponding to the center of the ERP image, which serves as the consistent reference viewpoint at longitude~0\degree{} and latitude~0\degree{}.}

\subsection{Image Collection}
To construct \ourbench{}, we collected 643 omnidirectional images from Flickr \cite{flick}, NOIRLab \cite{NOIRLab}, and Insta360~\cite{Insta360}. \trans{All images are in ERP format and publicly available for download. Figure~\ref{fig:category_distribution} shows the distribution across scene types and capture conditions. \ourbench{} is dominated by daytime, outdoor, urban scenes, but also includes indoor, drone/UAV, and nighttime images, reflecting diverse real-world scenarios.} The image resolutions range from $7296 \times 3648$ to $21344 \times 10672$; images larger than 7K were downsampled to $7296 \times 3648$ for consistency.


\subsection{Dataset Annotations}
\ot{Our 360Bench comprises 1,532 unique samples, each pairing a 360\degree{} image with a question and four answer options. All the questions and answer options are carefully designed by human annotators to ensure that correct answers cannot be inferred without truly understanding image content.}

Three trained annotators created the annotations following a detailed protocol. For each assigned 360\degree{} image, an annotator first explored the scene in Virtual Reality (VR) viewing mode, freely adjusting the viewing direction and zoom level to identify key objects. The objects were then assigned to subtasks based on their characteristics. For each sample, the annotator composed the corresponding question, along with its correct answer and three plausible yet incorrect answers. \trans{Bounding boxes were also annotated for the objects relevant to each question. 
Each annotator spent approximately three weeks completing their assigned set, reflecting the effort required to ensure high-quality annotations.} Quality control was enforced through cross-evaluation among annotators to resolve ambiguous cases. This procedure ensures a reliable and challenging benchmark for evaluating MLLMs on 360\degree{} image perception.

On average, each image contributes 2.38 unique samples (1,532 samples across 643 images). 
Each subtask contains from 213 to 225 unique samples, as shown in Fig.~\ref{fig:Breakdowntask}. To mitigate answer position bias, we applied cyclic permutation to the answer options~\cite{DC2,CycPer}, resulting in a total of 6,128 evaluation samples. 

Figure~\ref{fig:bbox_analysis_scatter} shows the spatial distribution of bounding boxes in \ourbench{}, where each point represents a box center. The annotations cover the full ERP image, with higher density near the equator, reflecting typical object layouts in 360° scenes.

\section{Proposed Method} \label{sec:Model}

\begin{figure*}[t]
\centering
\includegraphics[width=0.99\textwidth]{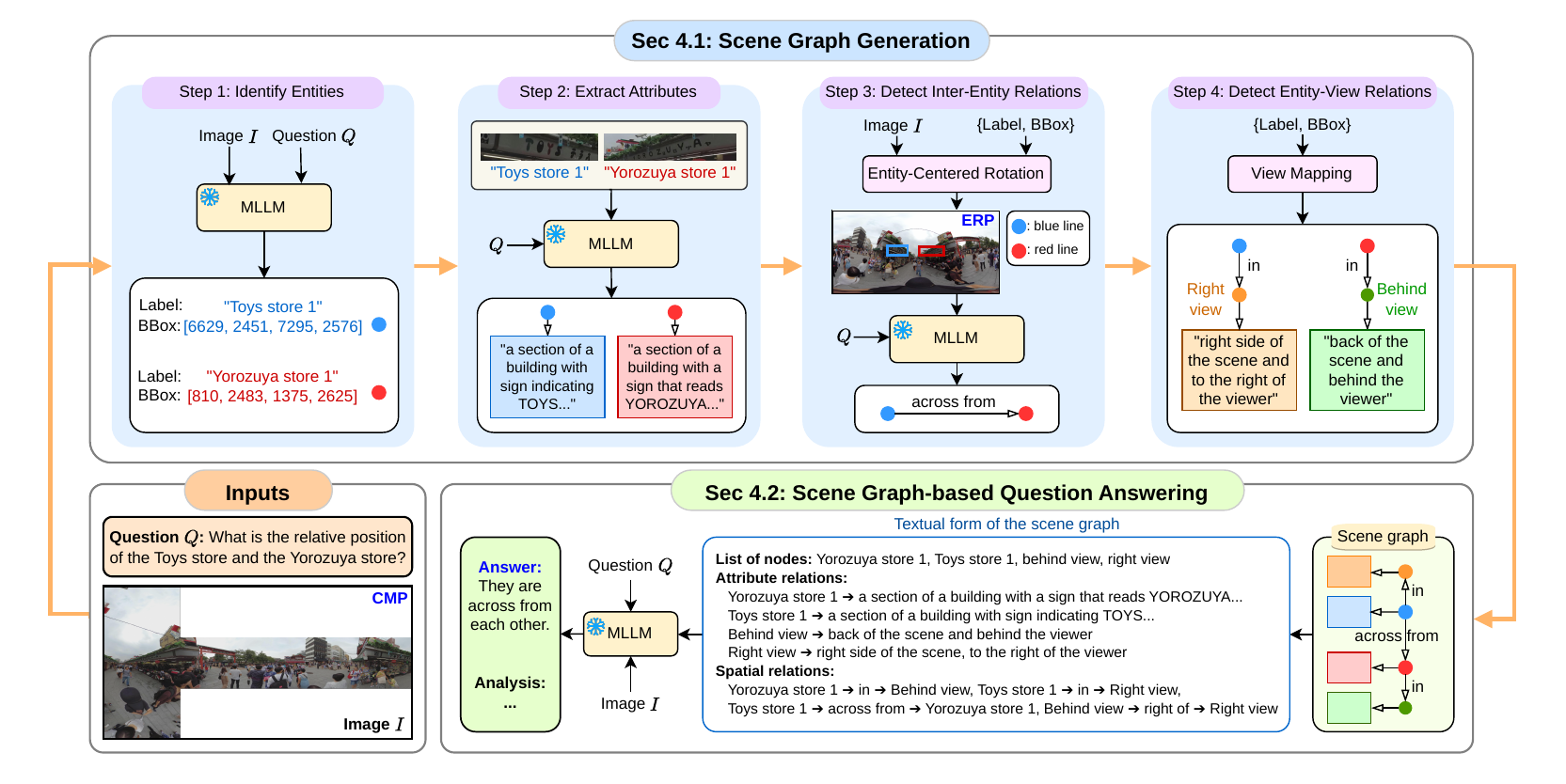} 
\caption{
\kj{Overview of \ourmodel{}. The model performs question answering via scene graph generation through four steps (Sec.~\ref{subsec:SGG}): 
(1) Entity Identification, detecting entities relevant to the question $Q$ from the CMP image $I$; 
(2) Attribute Extraction, deriving descriptive attributes from each entity crop; 
(3) Inter-Entity Relation Detection, capturing spatial relations among entities; and 
(4) \tran{Entity-View} Relation Detection, modeling spatial relations between entities and camera views. 
The resulting scene graph, represented in textual form, is then fed into the MLLM to generate the final answer and reasoning analysis (Sec.~\ref{subsec:answer}). 
360\degree{} image source: \cite{Insta360}.
}}

\label{fig:GeneralArchitecure}
\end{figure*}
\qt{Figure~\ref{fig:GeneralArchitecure} shows an overview of our proposed method, \textbf{Free360}. Given a question $Q$ and a 360\degree{} image $I$, \ourmodel{} answers the question by generating a question-relevant scene graph and reasoning over it. We decompose the SGG process into four explicit steps:
(1) {Entity Identification}, \ot{detecting} entities relevant to the question $Q$ from the image $I$;
(2) {Attribute Extraction}, \ot{deriving} descriptive attributes for each detected entity;
(3) {Inter-Entity Relation Detection}, \ot{capturing} spatial relations among entities \tran{based on spherically rotated images}; and
(4) {\tran{Entity-View} Relation Detection}, \ot{modeling} spatial relations between entities and camera views (or the viewer).}

Each step is guided by a task-specific prompt and a few in-context examples. 
\kj{Motivated by the complementary strengths of CMP and ERP observed in our analysis (Table~\ref{tab:perComparison}), we design Free360 as a hybrid framework that leverages CMP in Step 1 for robust object detection and ERP in Step 3 for spatial reasoning.} The resulting scene graph serves as an explicit reasoning scaffold, integrating pertinent information for the MLLM to infer the final answer.

\subsection{Scene Graph Generation} \label{subsec:SGG}

\ot{The goal of SGG is to construct a scene graph that captures the entities, their attributes, and spatial relations required for reasoning to answer the question $Q$.} 
\ot{We define a scene graph as $G = (\mathcal{N}, \mathcal{R})$, where $\mathcal{N}$ is the set of nodes and $\mathcal{R} \subseteq \mathcal{N} \times \mathcal{N}$ represents the sets of relations between them.}
\ot{Each node $n_i \in \mathcal{N}$ is defined as $n_i = (l_i, a_i)$, where $l_i$ and $a_i$ \qt{are} 
the node's label and attribute, respectively. Each directed relation $r_{i,j} \in \mathcal{R}$ represents a semantic relation from node $n_i$ to node $n_j$.}


\qt{The graph includes two types of nodes: (1) \textbf{entity nodes} $n^{e}$, representing \ot{visible 
entities within the image} relevant to the question, and (2) \textbf{view nodes} $n^{v}$, corresponding to the six cube faces of the CMP \ot{which represent the six principal directions relative to the viewer (i.e., top, bottom, left, right, front, behind)}. Accordingly, we define three types of relations: \textbf{inter-entity relations} $r^{ee}$, \textbf{entity-view relations} $r^{ev}$, and \textbf{inter-view relations} $r^{vv}$.}


Constructing $G$ with an MLLM is inherently challenging~\cite{PGSG,openworld}. To make this process tractable, we adopt a structured, step-by-step strategy inspired by the chain-of-thought paradigm~\cite{CoT,chen2025towards,zhao2025cot}, consisting of four main steps outlined below. 

\subsubsection*{Step 1: Entity Identification} \label{subsec:step1}


\qt{To alleviate geometric distortion, the CMP image $I$ is used as the visual input to \ourmodel{}. \ot{We prompt the MLLM with the CMP image $I$} and question $Q$ to identify and localize entities \ot{in the image that are} semantically relevant to the question. For each detected entity node $n_i^{e}$, \ot{the output contains} a label $l_i^{e}$ and 
its \ot{2D bounding box} $(x_{i,1}^{e}, y_{i,1}^{e}, x_{i,2}^{e}, y_{i,2}^{e})$, where $(x_{i,1}^{e}, y_{i,1}^{e})$ and $(x_{i,2}^{e}, y_{i,2}^{e})$ denote the top-left and bottom-right corners, respectively. \ot{Each entity is subsequently assigned a unique index to differentiate instances sharing the same label (e.g., `person 1')}.}

\subsubsection*{Step 2: Attribute Extraction} \label{subsec:step2}

\qt{For each entity node $n_i^{e}$, we extract its image crop from $I$ using the bounding box $(x_{i,1}^{e}, y_{i,1}^{e}, x_{i,2}^{e}, y_{i,2}^{e})$. Cropping \tran{not only reduces computational complexity but also} allows the model to focus on fine-grained features, such as text, patterns, or logos, that are difficult to discern from the full image. \ot{We prompt} the MLLM \ot{with the label and its associated image crop as inputs} to generate a textual attribute $a_i^{e}$ describing the entity in a manner relevant to answering the question $Q$. To integrate these attributes into the textual representation of the scene graph, we represent attribute relations as `$l_i^{e} \rightarrow a_i^{e}$', linking each entity node to its corresponding attribute.}

\subsubsection*{Step 3: Inter-Entity Relation Detection} \label{subsec:step3}

\qt{Reasoning about spatial relations in 360\degree{} scenes \ot{can} benefit from centering relevant entities, analogous to how humans shift their viewpoint. For each entity pair $(n_i^{e}, n_j^{e})$, we generate a rotated ERP image $I'$ that centers the pair, facilitating the MLLM's spatial reasoning. The rotation is parameterized by angles $\phi'$ (rotation about the $y$-axis) and $\theta'$ (rotation about the $z$-axis), computed from the bounding boxes $(x_{i,1}^{e}, y_{i,1}^{e}, x_{i,2}^{e}, y_{i,2}^{e})$ and $(x_{j,1}^{e}, y_{j,1}^{e}, x_{j,2}^{e}, y_{j,2}^{e})$.}

\qt{Specifically, we define the center $\mathbf{c}^*$ of the entity pair as the midpoint of the minimal spherical bounding box enclosing both entities. 
The rotation angles $(\phi', \theta')$ correspond to the longitude and latitude of $\mathbf{c}^*$.}
\ot{We obtain the rotated image $I'$ from $I$ by the rotation matrix}
\begin{equation*}  \scalemath{0.9}{
\mathbf{R}_\text{c} =
\begin{bmatrix}
\cos \phi' \cos \theta' & -\cos \phi' \sin \theta' & \sin \phi' \\
\sin \theta' & \cos \theta' & 0 \\
-\sin \phi' \cos \theta' & \sin \phi' \sin \theta' & \cos \phi'
\end{bmatrix}.}
\label{eq:rot_matrix1}
\end{equation*}
\ot{We then highlighted the entities in $I'$} with colored bounding boxes and a legend, i.e., ``\texttt{$l_i^e$:blue line,$l_j^e$:red line}''. \ot{We prompt the MLLM to analyze} $I'$ along with the legend to describe the spatial relation $r_{ij}^{ee}$ between the entities. Inter-entity relations are then encoded textually as `$l_i^{e} \rightarrow r_{ij}^{ee} \rightarrow l_j^{e}.$'

\begin{table}[t]
\centering
\caption{Labels and attributes of the six view nodes.}
\resizebox{0.75\columnwidth}{!}{
\begin{tabular}{p{2.6cm}l}
\toprule
\textbf{Label} & \textbf{Attribute} \\
\midrule
\texttt{left view} & Left side of the scene and to the left of the viewer. \\
\texttt{right view} & Right side of the scene and to the right of the viewer. \\
\texttt{front view} & Front of the scene and front of the viewer. \\
\texttt{behind view} & Back of the scene and behind the viewer. \\
\texttt{top view} & Top of the scene and above the viewer. \\
\texttt{bottom view} & Bottom of the scene and below the viewer. \\
\bottomrule
\end{tabular}}
\label{tab:view_attributes}
\end{table}

\subsubsection*{Step 4: Entity-View Relation Detection} \label{subsec:step4}
\qt{We define six view nodes $\mathcal{N}^{v} = \{n_q^{v}\}_{q=1}^{6}$ corresponding to the cube faces in the CMP format; see Table~\ref{tab:view_attributes}. Inter-view relations $r^{vv}$ are predefined and incorporated into the scene graph to capture spatial dependencies between views. For instance, the relation between \texttt{front view} and \texttt{behind view} is represented as `$\texttt{front view} \rightarrow \texttt{opposite} \rightarrow \texttt{behind view}$' \ot{since they lie on opposite cube faces.}}


\qt{To assign entities to views, we define a mapping function $f: \mathbb{R}^2 \rightarrow \mathcal{N}^{v}$ 
that maps 2D pixel coordinates in the CMP image to the corresponding view node. For each entity node $n_i^{e}$, we compute the center of its bounding box $(\bar{x}_i, \bar{y}_i)$,}
\qt{and determine its associated view node $n_q^{v} = f(\bar{x}_i, \bar{y}_i)$. The textual format of entity-view relation $r_{iq}^{ev}$ is then represented as `$l_i^{e} \rightarrow \texttt{in} \rightarrow l_q^{v}$',
linking each entity node to its corresponding view node in the scene graph.}

\subsection{Scene Graph-Based Question Answering} \label{subsec:answer}
\qt{The constructed scene graph from previous SGG steps is serialized into a structured textual form and fed to the MLLM for reasoning. Specifically, \ot{we represent a scene graph} with a list of relevant nodes (detected entities and views) and their spatial and attribute relations as follows:}  

\begin{tabular}
{
  >{\raggedleft\arraybackslash}p{0.3\linewidth}
  >{\raggedright\arraybackslash}p{0.4\linewidth}
}
\footnotesize
{List of nodes:} &
$\begin{cases}
\{l_i^{e}\}_{i=1}^{M}, \\
\{l_q^{v}\}_{q=1}^{K}
\end{cases}$ \\[1.5em]

\footnotesize
{Attribute relations:} &
$\begin{cases}
\{l_i^{e} \rightarrow a_i^{e}\}_{i=1}^{M}, \\
\{l_q^{v} \rightarrow a_q^{v}\}_{q=1}^{K}
\end{cases}$ \\[1.5em]

\footnotesize
{Spatial relations:} &
$\begin{cases}
\{l_i^{e} \rightarrow r_{iq}^{ev} \rightarrow l_q^{v}\}_{i=1}^{M}, \\
\{l_i^{e} \rightarrow r_{ij}^{ee} \rightarrow l_j^{e}\}_{1 \le i < j \le M}, \\
\{l_q^{v} \rightarrow r_{qp}^{vv} \rightarrow l_p^{v}\}_{1 \le q < p \le K}.
\end{cases}$
\vspace{4pt}
\end{tabular}

\begin{figure}[t]
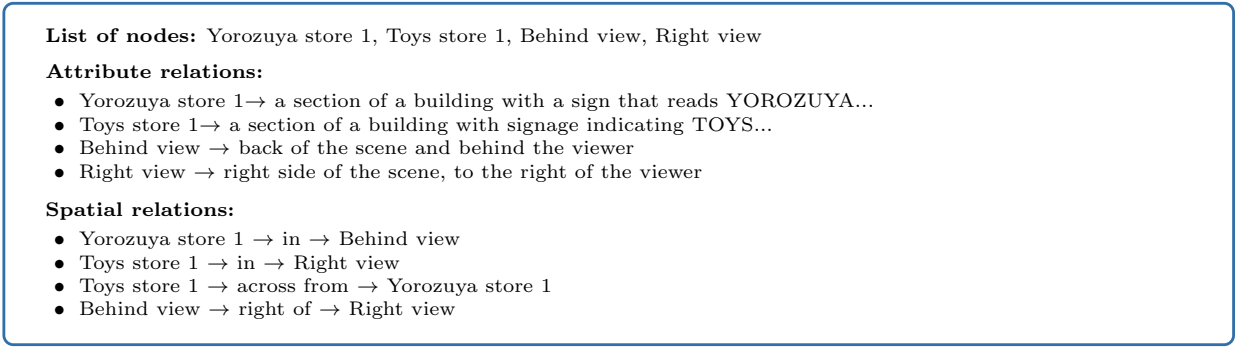

\centering
\resizebox{\textwidth}{!}{
\begin{tcolorbox}[
  width=0.93\linewidth,
  colback=white,
  colframe=navy,
  colbacktitle=navy,
  coltitle=white,
  fonttitle=\bfseries,
  arc=1.0mm,
  boxrule=1.pt
]
\scriptsize
\setlength{\parskip}{1.5pt}

\textbf{List of nodes:} Yorozuya store 1, Toys store 1, Behind view, Right view

\vspace{3pt}
\textbf{Attribute relations:}
\begin{itemize}[topsep=1pt, partopsep=0pt, parsep=0pt, itemsep=0.3pt, left=3pt]
\item Yorozuya store \ot{1}$\rightarrow$ a section of a building with a sign that reads YOROZUYA...
\item Toys store \ot{1}$\rightarrow$ a section of a building with signage indicating TOYS...
\item Behind view $\rightarrow$ back of the scene and behind the viewer
\item Right view $\rightarrow$ right side of the scene, to the right of the viewer
\end{itemize}

\vspace{3pt}
\textbf{Spatial relations:}
\begin{itemize}[topsep=1pt, partopsep=0pt, parsep=0pt, itemsep=0.3pt, left=3pt]
\item Yorozuya store 1 $\rightarrow$ in $\rightarrow$ Behind view
\item Toys store 1 $\rightarrow$ in $\rightarrow$ Right view
\item Toys store 1 $\rightarrow$ across from $\rightarrow$ Yorozuya store 1
\item Behind view $\rightarrow$ right of $\rightarrow$ Right view
\end{itemize}

\end{tcolorbox}}
\caption{\qt{Example of a scene graph in the serialized textual form.}}
\label{fig:scene_graph_example}
\end{figure}

\noindent
\qt{Here, $M$ and $K$ denote the numbers of entities and views, respectively. {Figure~\ref{fig:scene_graph_example} shows a serialized scene graph example.}}

\qt{To generate an answer, we prompt the MLLM with instructions to produce the reasoning analysis and select the most plausible option, along with the serialized graph text, the question, and its answer options. To ensure answer reliability, the MLLM is instructed to output `\texttt{CANNOT ANSWER}' if the provided graph lacks sufficient information. 
In such cases, the image $I$ is used as a fallback input to the MLLM for generating the final answer.}




\section{Experiments}\label{sec:Experiments}

\subsection{Human Evaluation}


\tran{We conducted a subjective study to evaluate human performance on \ourbench{}. Five participants (university students and staff aged 22--34) took part in the study. 
\qt{We developed a web-based user interface using A-Frame~\cite{aframevr2025} to present 360\degree{} images in VR mode to the participants.}
\qt{The} participants could freely adjust their viewing direction to explore each 360° scene using a mouse. 
To mitigate potential motion sickness from prolonged 360\degree{} viewing, each participant answered 70 randomly sampled questions (10 per subtask) from \ourbench{}, taking approximately 20--50 minutes.}


\subsection{Experimental Settings} 
\noindent
\textbf{Evaluated models.}  \qt{We evaluate seven representative open-source and proprietary MLLMs on \ourbench{}:}
\qt{GPT-4o~\cite{GPT4o}, Gemini Flash 2.5~\cite{gemini25}, Gemini Pro 2.5~\cite{gemini25}, Qwen2.5-VL-7B~\cite{qwen25vltechnicalreport}, DeepSeek-VL2-Small~\cite{deepseekvl2}, 
LLaVA-v1.5-7B~\cite{llavav15}, and LLaVA-CoT~\cite{llavacot}.}
We also evaluate six enhancement methods:  SEAL~\cite{SEAL}, DC$^2$~\cite{DC2}, ZoomEye~\cite{zoomeye}, VisCoT-7B-336~\cite{VisualCoT}, DeepEyes~\cite{deepeyes}, and Omni-CoT~\cite{odibench}. 

\qt{Although several other 360\degree{} VQA models have been proposed~\cite{VQA360,omniVQA}, they are excluded from our evaluation because (1) their required large-scale training datasets are unavailable, (2) their pretrained weights have not been released, and (3) models such as~\cite{VQA360} are not compatible with MLLM-based inference.}

\noindent
\textbf{Evaluation \tran{metrics}.} \qt{We report accuracy as the primary evaluation metric, calculated as the proportion of correctly answered samples. \ot{Except for proprietary models whose inference time is measured via API calls, we also report the average per-sample inference time on an \trans{NVIDIA H200 GPU} (in seconds) to assess computational efficiency.} 
}

\noindent
\textbf{Implementation details.} 
All models are configured to generate outputs deterministically using greedy decoding to ensure reproducibility. 
We convert ERP images into CMP format using 360Lib \cite{360lib}. The CMP resolution is set to $7296\times5472$, matching the equatorial pixel count of the ERP format to ensure a fair comparison. 
Regarding image processing, proprietary models directly handle images via their APIs. \trans{For other models, images are preprocessed according to their default settings to ensure evaluation under their intended operating conditions.} Specifically, Qwen2.5-VL-7B and its enhancement methods process ERP images at $5068\times2520$ and CMP images at $4116\times3080$. DeepSeek-VL2-Small uses dynamic tiling, with tile sizes of 384 and 1024. The remaining models resize inputs to their maximum supported resolutions---224 for SEAL and 336 for the others.  

For the enhancement methods, SEAL, VisCoT-7B-336, and DeepEyes are tested using their official fine-tuned models, while the training-free methods DC$^2$, ZoomEye, and Omni-CoT adopt Qwen2.5-VL-7B as the base MLLM. \trans{Since Omni-CoT's source code is unavailable at the time of writing, we re-implement it based on the original paper.} For a fair comparison with other training-free methods, we also employ Qwen2.5-VL-7B as its base model for \ourmodel{}.

\subsection{
\qt{Results on \ourbench{}}}

\qt{Table~\ref{tab:perComparison} compares human and model performance on our \ourbench{}. Human participants achieve the highest accuracy of 86.3\% by immersing themselves in 360\degree{} images and answering each question in 28.9 seconds on average. They perform strongly on recognition, spatial reasoning, and direction-giving tasks ($\ge$84\%) but show lower accuracy on counting (82\% for FP-IC and 78\% for PP-IC), highlighting the inherent counting difficulty in complex 360\degree{} scenes.}
 

State-of-the-art MLLMs, including GPT-4o and Gemini Pro 2.5, still struggle with omnidirectional perception. While Gemini Pro 2.5 achieves the best overall accuracy (46.5\%), it falls far short of human-level accuracy (86.3\%). Furthermore, its black-box nature makes direct comparisons with open-source models less informative. Among open-source models, Qwen2.5-VL-7B leads with 38.1\%, underscoring the persistent gap between current MLLMs and humans.

Projection format strongly influences MLLM's performance. Overall, CMP achieves higher accuracy than ERP across most models (up to +2.6\% for Gemini Pro 2.5). CMP shows particularly strong gains on projection-distorted tasks (e.g., +14.1\% on PP-IR with LLaVA-CoT and +12.2\% on PP-IC with Gemini 2.5 Flash). In contrast, ERP generally performs better on spatial reasoning tasks (e.g., +2.4\% on SR-Os for LLaVA-CoT and \kj{+14.5\%} on SR-OV for Gemini 2.5 Flash, and +6.8\% on DG for Gemini 2.5 Pro). These results reveal complementary strengths between the two formats: \trans{CMP is more effective for object perception under projection distortion, whereas ERP benefits spatial reasoning.}

\begin{table*}[t]
  \renewcommand{\arraystretch}{1.12}
  \setlength\tabcolsep{2pt}
  \centering
  \caption{Human and model evaluation results. The highest score for each task is highlighted in bold. \textit{Acc.} denotes per-task accuracy; \textit{Difference} indicates the difference between \ourmodel{} and Qwen2.5-VL-7B in accuracy and inference time.}
  \label{tab:perComparison}
  \resizebox{1.0\textwidth}{!}{
  \begin{tabular}{l|c|ccc|ccc|ccc|c|c|c}
  \Xhline{1.2pt}
  \multirow{2}{*}{Model} & \multirow{2}{*}{\begin{tabular}[c]{@{}c@{}}Projection\\ format\end{tabular}} & \multicolumn{3}{c|}{FP (Acc.)} & \multicolumn{3}{c|}{PP (Acc.)} & \multicolumn{3}{c|}{SR (Acc.)} & {DG} & {Overall} & Inf. Time \\
  \cline{3-11}
  &  & IR & IC & Avg. & IR & IC & Avg. & Os & OV & Avg. & (Acc.) & (Acc.) & (seconds) \\
   \hline \hline
  Random guess & --- & 25.0 & 25.0 & 25.0 & 25.0 & 25.0 & 25.0 & 25.0 & 25.0 & 25.0 & 25.0 & 25.0 & --- \\
  \hline
  \textbf{Human} & --- & 96.0 & 82.0 & 89.0 & 86.0 & 78.0 & 82.0 & 84.0 & 90.0 & 87.0 & 88.0 & 86.3 & 28.9 \\
  \hline \hline
  \rowcolor{gray!10} \multicolumn{14}{c}{\textbf{MLLMs}} \\
  \hline
  \multirow{2}{*}{GPT-4o}	&	ERP	&	48.6	&	32.0	&	40.3	&	54.2	&	38.8	&	46.5	&	37.7	&	35.7	&	36.7	&	38.5	&	40.7	&	1.5	\\  
	&	CMP	&	45.5	&	30.8	&	38.1	&	\textbf{63.0}	&	44.0	&	\textbf{53.5}	&	37.4	&	34.0	&	35.7	&	35.3	&	41.3	&	1.6	\\   \hline
\multirow{2}{*}{Gemini Flash 2.5} 	&	ERP	&	62.3	&	40.0	&	51.0	&	44.5	&	33.8	&	39.2	&	37.3	&	40.3	&	38.8	&	40.6	&	42.7	&	10.8	\\  
	&	CMP	&	57.5	&	41.9	&	49.6	&	56.5	&	46.0	&	51.3	&	37.1	&	25.8	&	31.5	&	34.7	&	42.6	&	12.5	\\   \hline
 \multirow{2}{*}{Gemini Pro 2.5}	&	ERP	&	\textbf{63.2}	&	43.5	&	\textbf{53.3}	&	41.1	&	34.7	&	37.9	&	38.0	&	\textbf{42.9}	&	40.5	&	\textbf{44.0}	&	43.9	&	18.4	\\  
	&	CMP	&	61.5	&	\textbf{44.8}	&	53.0	&	52.2	&	\textbf{46.7}	&	49.5	&	\textbf{46.8}	&	36.9	&	\textbf{41.9}	&	37.2	&	\textbf{46.5}	&	18.3	\\   \hline
\multirow{2}{*}{Qwen2.5-VL-7B}	&	ERP	&	59.5	&	27.9	&	43.6	&	42.2	&	36.0	&	39.1	&	30.7	&	31.2	&	31.0	&	33.7	&	37.3	&	2.1	\\  
	&	CMP	&	58.1	&	27.0	&	42.4	&	50.6	&	42.0	&	46.3	&	28.4	&	30.4	&	29.4	&	30.6	&	38.1	&	2.2	\\   \hline
\multirow{2}{*}{DeepSeek-VL2-Small}	&	ERP	&	41.8	&	22.7	&	32.2	&	42.4	&	39.3	&	40.9	&	27.2	&	32.6	&	29.9	&	32.4	&	33.9	&	0.9	\\  
	&	CMP	&	40.3	&	19.2	&	29.7	&	55.4	&	43.1	&	49.2	&	28.2	&	29.7	&	28.9	&	31.1	&	35.1	&	0.9	\\   \hline
\multirow{2}{*}{LLaVA-v1.5-7B}	&	ERP	&	30.7	&	8.4	&	19.4	&	29.3	&	35.9	&	32.6	&	22.4	&	26.7	&	24.5	&	26.6	&	25.6	&	\textbf{0.4}	\\  
	&	CMP	&	32.5	&	9.3	&	20.8	&	33.4	&	39.6	&	36.5	&	23.5	&	23.9	&	23.7	&	26.3	&	26.8	&	0.5	\\   \hline
\multirow{2}{*}{LLaVA-CoT}	&	ERP	&	34.9	&	16.7	&	25.7	&	41.5	&	33.3	&	37.4	&	27.2	&	30.0	&	28.6	&	27.5	&	30.1	&	2.7	\\  
	&	CMP	&	33.3	&	16.9	&	25.0	&	55.6	&	37.3	&	46.5	&	24.8	&	30.9	&	27.8	&	27.6	&	32.2	&	3.0	\\   \hline
\hline    \rowcolor{blue!10} \multicolumn{14}{c}{\textbf{MLLM Enhancement Models}}  \\ \hline																											
\multirow{2}{*}{SEAL}	&	ERP	&	31.3	&	24.6	&	27.9	&	30.4	&	27.8	&	29.1	&	23.6	&	20.4	&	22.0	&	18.9	&	25.2	&	14.6	\\  
	&	CMP	&	33.6	&	21.4	&	27.5	&	32.1	&	30.4	&	31.3	&	23.6	&	18.7	&	21.1	&	20.7	&	25.7	&	14.6	\\   \hline
\multirow{2}{*}{DC$^2$}	&	ERP	&	40.0	&	16.0	&	27.9	&	44.0	&	36.4	&	40.2	&	22.0	&	25.1	&	23.6	&	21.5	&	29.1	&	617.9	\\  
	&	CMP	&	39.7	&	11.0	&	25.2	&	49.1	&	34.9	&	42.0	&	20.2	&	23.1	&	21.6	&	20.2	&	28.1	&	761.6	\\   \hline
\multirow{2}{*}{ZoomEye}	&	ERP	&	\textbf{61.5}	&	26.6	&	43.9	&	44.2	&	31.9	&	38.1	&	25.3	&	31.3	&	28.3	&	28.5	&	35.5	&	19.5	\\  
	&	CMP	&	59.2	&	22.8	&	40.8	&	51.2	&	30.3	&	40.7	&	27.3	&	29.1	&	28.2	&	28.0	&	35.3	&	23.9	\\   \hline
\multirow{2}{*}{VisCoT-7B-336}	&	ERP	&	28.1	&	23.3	&	25.7	&	30.6	&	35.9	&	33.3	&	22.3	&	26.7	&	24.4	&	23.8	&	27.2	&	1.1	\\  
	&	CMP	&	28.3	&	23.7	&	26.0	&	34.8	&	36.4	&	35.6	&	24.6	&	24.4	&	24.5	&	24.2	&	28.0	&	\textbf{0.8}	\\   \hline
\multirow{2}{*}{DeepEyes}	&	ERP	&	53.6	&	25.2	&	39.3	&	33.3	&	32.0	&	32.7	&	28.2	&	26.9	&	27.5	&	29.1	&	32.6	&	78.4	\\  
	&	CMP	&	52.6	&	25.6	&	39.0	&	43.5	&	42.5	&	43.0	&	31.4	&	26.3	&	28.9	&	27.3	&	35.5	&	66.9	\\   \hline
Omni-CoT	&	Hybrid	&	58.9	&	24.3	&	41.4	&	49.5	&	40.3	&	44.9	&	28.3	&	41.0	&	34.7	&	35.0	&	39.5	&	16.6	\\   \hline
\rowcolor{blue!5} \textbf{Free360 (Ours)}	&	Hybrid	&	60.7	&	\textbf{30.0}	&	\textbf{45.2}	&	\textbf{54.7}	&	\textbf{43.9}	&	\textbf{49.3}	&	\textbf{33.6}	&	\textbf{53.3}	&	\textbf{43.6}	&	\textbf{41.4}	&	\textbf{45.3}	&	22.5	\\  
{\textit{Difference} w.r.t}	&	ERP	&\imt{+	1.1	}&\imt{+	2.1	}&\imt{+	1.6	}&\imt{+	12.5	}&\imt{+	7.9	}&\imt{+	10.2	}&\imt{+	3.0	}&\imt{+	22.1	}&\imt{+	12.6	}&\imt{+	7.8	}&\imt{+	8.1	}&{\color{timecolor}+	20.3	}\\  
Qwen2.5-VL-7B	&	CMP	&\imt{+	2.6	}&\imt{+	3.0	}&\imt{+	2.8	}&\imt{+	4.1	}&\imt{+	1.9	}&\imt{+	3.0	}&\imt{+	5.2	}&\imt{+	22.9	}&\imt{+	14.2	}&\imt{+	10.9	}&\imt{+	7.3	}&{\color{timecolor}+	18.1	}\\  \Xhline{1pt}
  \end{tabular}
  }
  \end{table*}

Among enhancement methods, ZoomEye improves fine-grained perception on FP-IR (+1.9\%) but weakens holistic reasoning on FR-IC \kj{(-1.3\%)} 
over its base MLLM (Qwen2.5-VL-7B), indicating that its patch-wise search favors local detail over global understanding. 
\trans{For Omni-CoT, gains of +10.6\% on SR-OV and +4.4\% on DG suggest that providing explicit view-level information effectively strengthens viewer-centric spatial reasoning. However, Omni-CoT shows marginal declines on most other tasks (up to -2.4\%), implying that relying solely on view-level decomposition and object-level cropping may be insufficient for more complex and holistic reasoning.}


\trans{In contrast, \textbf{\ourmodel{}} achieves the best overall accuracy of 45.3\%, outperforming Qwen2.5-VL-7B with CMP by +7.3\% and consistently improving performance across all tasks (up to +22.9\% on SR-OV). These results demonstrate that scene graph-based reasoning, empowered by 360°-specific operations, enables more robust and comprehensive understanding in omnidirectional settings.} These improvements are achieved with a moderate increase in inference time (from 2.1 to 22.5 seconds). \trans{Nevertheless, the resulting latency is still within the range of human response time (28.9 seconds). It is comparable to ZoomEye (19.5--23.9 seconds) and slightly higher than Omni-CoT (16.6 seconds), while being substantially faster than DeepEyes (66.9--78.4 seconds) and DC$^2$ (617.9--761.6 seconds).}


\subsection{Ablation Study}

\begin{table}[t]
\centering
\setlength{\tabcolsep}{8pt}
\renewcommand{\arraystretch}{1.1}
\caption{\qt{
Ablation study of \ourmodel{}. 
}
\label{tab:ablation}
}
\resizebox{\textwidth}{!}{
\begin{tabular}{@{}c@{  ~~~~  }c@{}}
\begin{tabular}{l|cc}
\multicolumn{3}{c}{(a) Impact of SGG components} \\
\Xhline{1pt}
\textbf{Setting} & \textbf{Acc.} & \textbf{Inf. Time} \\
\hline
\rowcolor{blue!5} Full       & \textbf{45.3} & 22.5 \\
\rowcolor{white}  w/o Crop   & 42.7          & 23.8 \\
w/o Rotate                  & 44.7          & \textbf{21.0} \\
w/o EVR                     & 42.8          & 21.1 \\
\Xhline{1pt}
\end{tabular}
&
\begin{tabular}{c|cc|c|cc}
\multicolumn{6}{c}{(b) Effectiveness across model scales} \\
\Xhline{1pt}
\textbf{Model}	&	\multicolumn{2}{c|}{\textbf{Qwen2.5-VL}}	&			\textbf{Free360}	&	\multicolumn{2}{c}{\textbf{Acc. Gain}}			\\\cline{2-6}
\textbf{scale}	&	\textit{ERP}	&	CMP	&	Hybrid	&	ERP	&	CMP	\\ \hline
3B	&	35.2	&	37.2	&	40.8	&\imt{+	5.5	}&\imt{+	3.5	}\\
\rowcolor{blue!5}7B	&	37.3	&	38.1	&	\textbf{45.3}	&\textbf{\imt{+	8.1	}}&\textbf{\imt{+	7.3	}}\\
\rowcolor{white} 32B	&	\textbf{37.8}	&	\textbf{39.7}	&	45.0	&\imt{+	7.2	}&\imt{+	5.3	}\\
\Xhline{1pt}
\end{tabular}
\end{tabular}
}
\end{table}

We conduct an ablation study to evaluate the contribution of each component in \ourmodel{} and its effectiveness across different model scales, as summarized in Table~\ref{tab:ablation}.
Specifically, \textit{Full} denotes \qt{the complete model}; \textit{w/o Crop} uses the full image instead of image crops for the attribute extraction; \textit{w/o Rotate} {denotes removing} the spherical rotation in the inter-entity relation detection; and \textit{w/o EVR} excludes the entity-view relation detection step. Under the \textit{Full} setting, we further assess the effectiveness of Free360 across varying base MLLM scales (i.e., 3B, 7B, and 32B).

\qt{Table~\ref{tab:ablation}(a) isolates the contribution of each design choice in \ourmodel{}. Using image cropping for attribute extraction yields a 2.7\% accuracy gain and reduces inference time by 1.4~seconds, suggesting improved focus and efficiency. Spherical rotation enhances inter-entity relation detection, adding 0.7\% accuracy at a 1.5~second inference cost. Finally, modeling entity-view relations provides a further 2.6\% improvement with an overhead of 1.4~seconds, confirming its complementary benefit.}

\trans{Table~\ref{tab:ablation}(b) reports the performance of the base MLLM (Qwen2.5-VL) and our Free360 under the Full setting across different model scales. Free360 consistently improves performance across all base model sizes. The largest gain is observed at 7B, where Free360 achieves 45.3\% accuracy, yielding improvements of +8.1\% over ERP and +7.3\% over CMP. For 3B and 32B, the improvements range from +3.5\% to +7.2\%. These results confirm the general effectiveness of Free360 across varying model scales.}


\section{Conclusion and Future Work}\label{sec:Conclusion}
In this paper, we have presented a comprehensive study on 360\degree{} image perception with MLLMs through \ourbench{}, a 360\degree{} VQA benchmark. Using \ourbench{}, we have systematically evaluated thirteen state-of-the-art models, including both MLLMs and enhanced variants, revealing their limitations in reasoning over panoramic scenes. To address this, we have introduced \ourmodel{}, a training-free, scene-graph-based framework for high-resolution 360\degree{} VQA, empowered by 360-specific operations, including entity-centered rotation and view mapping. \ourmodel{} consistently improves its base MLLM across all tasks, achieving a 7.3\% overall improvement while remaining computationally efficient.

\qt{Future work on training-free methods using scene graphs for 360\degree{} tasks may explore finer perception modules to strengthen local reasoning, such as patch-wise retrieval. Extending \ourmodel{} for 360\degree{} video understanding is also a promising direction. We hope this study provides a foundation for advancing multimodal reasoning in omnidirectional environments and inspires further development of robust, generalizable MLLMs for 360\degree{} perception.}

%
%
\bibliographystyle{splncs04}
\bibliography{main}
\end{document}